\documentclass{article} 
\usepackage{iclr2027_conference,times}

\usepackage{amsmath,amsfonts,bm}

\def\eqref#1{equation~\ref{#1}}

\def\1{\bm{1}}

\DeclareMathAlphabet{\mathsfit}{\encodingdefault}{\sfdefault}{m}{sl}
\SetMathAlphabet{\mathsfit}{bold}{\encodingdefault}{\sfdefault}{bx}{n}

\iclrfinalcopy 
\usepackage{url}
\usepackage{subcaption}
\usepackage{float}
\usepackage{makecell}
\usepackage{graphicx}
\usepackage{wrapfig}
\usepackage{booktabs}
\usepackage{amssymb}
\usepackage[dvipsnames,svgnames]{xcolor}
\usepackage{listings}
\usepackage[T1]{fontenc}

\usepackage[ruled,vlined,linesnumbered]{algorithm2e}
\usepackage{hyperref}
\usepackage{cleveref}
\AtBeginDocument{%
  }

\usepackage[english]{babel}
\addto\extrasenglish{}
\addto\extrasenglish{}

\definecolor{linkcolor}{HTML}{647382}
\definecolor{citecolor}{HTML}{647382} %
\definecolor{urlcolor}{rgb}{0.4,0.2,0.2}
\definecolor{sqlcolor}{HTML}{965d67}
\definecolor{smtcolor}{HTML}{5d968c}
\definecolor{webblue}{rgb}{0,0,.7}
\definecolor{webgreen}{rgb}{0,.5,0}
\definecolor{webbrown}{rgb}{.6,0,0}
\definecolor{notecolor}{HTML}{FFF8DC}

\usepackage{textcomp}

\usepackage{xcolor}

\usepackage{float}
\newfloat{listing}{ht}{lol}
\floatname{listing}{Listing}

\usepackage{xspace}
\usepackage{fvextra}

\newcommand{\sys}{\mbox{\textsc{ActionEngine}}\xspace}

\newcommand{\plan}{Planner\xspace}
\newcommand{\exec}{Executor\xspace}
\newcommand{\patch}{Patcher\xspace}

\newcommand{\eg}{\emph{e.g.,}\xspace}

\newcommand{\minihead}[1]{{\vspace{.45em}\noindent\textbf{#1.}}}

\newcommand{\squishitemize}{
 \begin{list}{$\bullet$}
  { \setlength{\itemsep}{0pt}
     \setlength{\parsep}{0pt}
     \setlength{\topsep}{0pt}
     \setlength{\partopsep}{0pt}
     \setlength{\leftmargin}{1.95em}
     \setlength{\labelwidth}{1.5em}
     \setlength{\labelsep}{0.5em} } }

\newcounter{Lcount}
\newcommand{\squishlist}{
    \begin{list}{\arabic{Lcount}. }
   { \usecounter{Lcount}
        \setlength{\itemsep}{0pt}
        \setlength{\parsep}{3pt}
        \setlength{\topsep}{0pt}
        \setlength{\partopsep}{0pt}
        \setlength{\leftmargin}{2em}
        \setlength{\labelwidth}{1.5em}
        \setlength{\labelsep}{0.5em} } }

\newcommand{\squishend}{\end{list}}

\lstdefinestyle{PyStyle}{
  language=Python,
  basicstyle=\ttfamily\scriptsize, 
  aboveskip = 0.05in,
  belowskip = 0.05in,
  breaklines=true,
  frame=none,
  numbers=left,
  numberstyle=\tiny\color{gray},
  numbersep=8pt,
  stepnumber=1,
  captionpos=b,
  showstringspaces=false,
  tabsize=4,
  keepspaces=true,
  keywordstyle=\color{blue}\bfseries,
  stringstyle=\color{red!70!black},
  commentstyle=\color{green!60!black}\itshape,
  identifierstyle=\color{black},
  emphstyle=\color{purple}\bfseries,
  morekeywords={self, def, for, sum, in, and, async, await, dict, isinstance, enumerate, range, len},
  emph={__init__, __str__, __repr__},
  escapechar=|,
  morestring=[b]',
  morestring=[b]"  
}

\lstdefinestyle{YAMLStyle}{
  basicstyle=\ttfamily\scriptsize,
  aboveskip=6pt,
  belowskip=6pt,
  breaklines=true,
  breakatwhitespace=true,
  xleftmargin=15pt,
  xrightmargin=10pt,
  rulecolor=\color{black!20},
  numbers=left,
  numberstyle=\tiny\color{gray!60},
  numbersep=8pt,
  stepnumber=1,
  captionpos=b,
  showstringspaces=false,
  tabsize=2,
  keywordstyle=\color{blue!80!black}\bfseries,
  stringstyle=\color{red!70!black},
  commentstyle=\color{green!60!black}\itshape,
  emphstyle=\color{purple!80!black}\bfseries,
  morecomment=[l]{\#},
  morestring=[b]',
  morestring=[b]",
  keywords={name, src_state, dst_state, actions, type, input, locator, selector, output_format, ui_elements, atoms, index, description, action, locator_all},
  emph={Operation, Atom, State, click, read_text_all, fill, inner_text, inner_text_all},
  literate=
    {-}{{\textcolor{blue!60!black}{-}}}1
    {:}{{\textcolor{black!60}{:}}}1
    {&}{{\textcolor{orange!80!black}{\&}}}1
    {!}{{\textcolor{orange!80!black}{!}}}1
}

\lstdefinelanguage{json}{
  string=[b]",
  stringstyle=\color{red!70!black},
  comment=[l]{//},
  morecomment=[s]{/*}{*/},
  literate=
    {:}{{{\color{black!60}{:}}}}{1}
    {,}{{{\color{black!60}{,}}}}{1}
}

\lstdefinestyle{JSONStyle}{
  language=json,
  basicstyle=\ttfamily\scriptsize,
  aboveskip=6pt,
  belowskip=6pt,
  breaklines=true,
  breakatwhitespace=true,
  xleftmargin=15pt,
  xrightmargin=10pt,
  frame=single,
  frameround=tttt,
  framesep=4pt,
  rulecolor=\color{black!20},
  numbers=left,
  numberstyle=\tiny\color{gray!60},
  numbersep=8pt,
  stepnumber=1,
  captionpos=b,
  showstringspaces=false,
  tabsize=2
}

\begin{document}


\title{\sys: From Reactive to Programmatic Web Agents via State Machine Memory}


\author{Hongbin Zhong$^{\dagger}$,
Fazle Faisal$^{\star}$,
Luis Fran\c{c}a$^{\star}$,
Tanakorn Leesatapornwongsa$^{\star}$, \\
\textbf{Adriana Szekeres}$^{\star}$,
\textbf{Kexin Rong}$^{\dagger}$,
\textbf{Suman Nath}$^{\star}$ \\
Georgia Tech$^{\dagger}$,
Microsoft$^{\star}$ \\
\texttt{\{hzhong81,krong\}@gatech.edu},\\
\texttt{\{fafaisal,luis.franca,taleesat,aszekeres,sumann\}@microsoft.com}
}




\maketitle
\pagestyle{plain}

\begin{abstract}
Many web agents operate through a reactive execution loop: they observe the current interface, reason about the next action, execute it, and repeat.
This design incurs latency and cost that grow with the number of actions, while requiring agents to repeatedly rediscover how the same web application works.
We present \sys, a novel architecture that replaces step-by-step reasoning with programmatic execution using reusable knowledge of the application.
A Crawling Agent explores the application offline and constructs an updatable
\emph{state-machine memory} that represents its interface states, the operations available in each state, and the transitions between states.
Unlike trajectory memory, this representation stores how the application works rather than solutions to individual tasks.
At runtime, an Execution Agent uses this memory to synthesize a complete executable program in a single planning step, which is then executed deterministically without further planning calls.
When the interface changes or the memory is incomplete, a reactive fallback repairs the failed action and updates the memory for future tasks.
On 655 tasks across four WebArena domains, \sys achieves a 91.2\% success rate, outperforming the strongest reactive baseline, Claude Code, by 8.5 percentage points while reducing average task latency by $3.2\times$ and cost by $8\times$ using the same backbone model.

\end{abstract}







\section{Introduction}

LLM-based web agents automate browser tasks by observing web pages and performing actions such as clicking, typing, and extracting information~\cite{nguyen2025guiagentssurvey}.
Most web agents follow a reactive execution paradigm: the agent observes the current page, reasons about the next action, executes it, and repeats~\cite{yao2022react}.
This is a natural approach to an unfamiliar task or environment, where the agent must explore and adapt to what it observes, much like a user encountering an application for the first time.

However, web applications are not entirely new each time they are used.
Tasks within the same application share a relatively stable structure: the same pages, controls, operations, and navigation paths recur across tasks.
Reactive agents do not fully exploit this structure.
They repeatedly rediscovers how the application works, invoking the LLM throughout execution.
As a result, their cost and latency grow with the number of actions, and errors can accumulate over long trajectories~\cite{cheng2024seeclick, tao2023webwise, shinn2023reflexion, he2024webvoyager}.

Prior work reuses experience by storing demonstrations or trajectories from previous executions~\cite{zheng2023synapse,gu2024webdreamer,chae2024web}.
However, a trajectory packages together the decisions made for one particular task; a new task may reuse only some of its actions, possibly in a different order or context.
We argue that representing the \emph{application itself} offers a more compositional form of reuse.
For example, knowledge of how to search for a product, read its reviews, and edit its description can be recombined into workflows that were never previously executed.
This leads to our central question:
\emph{Can web agents replace step-by-step reactive reasoning with offline application knowledge and one-shot planning?}

\begin{figure*}[ht]
    \centering
    \includegraphics[width=\textwidth]{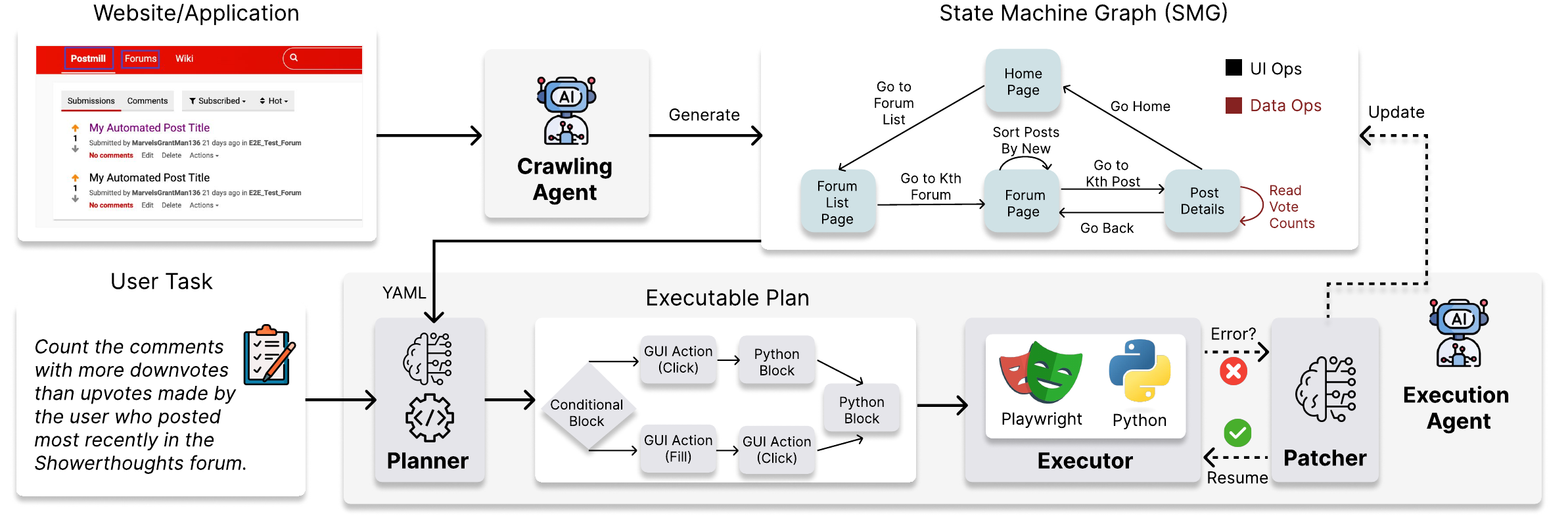} 
\vspace{-1em}
\caption{Overview of \sys. The \textbf{Crawling Agent} (top, offline) systematically explores the website to construct the state-machine graph (SMG). 
The \textbf{Execution Agent} (bottom, online) uses the SMG to plan and
execute tasks: the \emph{\plan} synthesizes an executable program from
the SMG and user task; the \emph{\exec} runs it deterministically
without further planning calls; and the \emph{\patch} intercepts
failures, invokes a reactive fallback to recover, and updates the SMG.
}
\vspace{-1.5em}
\label{fig:architecture}
\end{figure*}

We present \sys, an architecture that moves application discovery offline and uses the resulting knowledge to plan and execute new tasks programmatically (\autoref{fig:architecture}).
A Crawling Agent explores a web application and constructs an updatable
\emph{state-machine memory}.
The memory represents the application's interface states, the operations available in each state, and the transitions between states. 
To keep this memory representation compact, \sys separates page templates from dynamic content: for example, different product pages are represented as instances of the same underlying state.
Unlike trajectory memory, state-machine memory captures task-independent knowledge of how the application works, allowing operations learned from different interactions to be composed into programs for new tasks.

At runtime, an Execution Agent uses this memory to synthesize a high-level program sketch in a single LLM call.
A deterministic compiler then grounds each step of the sketch against operations in the state-machine memory to produce an executable plan, which runs without further model calls.
In the common case, this changes the unit of LLM reasoning from one call per action to one call per task.
Planning against an application-level model also gives the agent access to states and transitions beyond its current observation, rather than requiring each decision to be made locally from the current page.
When the memory is incomplete, the interface has changed, or execution depends on information that could not be anticipated, \sys falls back to reactive reasoning.
Successful repairs to stale or incorrect application knowledge are incorporated back into the memory for future tasks.

We evaluate \sys on 655 tasks across four WebArena domains~\cite{zhou2023webarena}.
Using the same backbone model, \sys achieves a 91.2\% success rate, 8.5 percentage points higher than the strongest reactive baseline, Claude Code, while reducing average task latency by $3.2\times$ and query-time cost by $8\times$.
The offline construction cost is amortized over repeated use: across the evaluated applications, \sys recovers the one-time crawling cost after 39--101 tasks through lower per-task execution cost.
Our analysis also reveals the boundary of programmatic execution: tasks that depend on unpredictable or noisy search results remain difficult to plan completely in advance and often benefit from reactive observation and adaptation.

Our contributions are threefold.
First, we introduce \emph{state-machine memory}, a task-independent and compositional representation of an application's interface states, operations, and transitions that allows application discovery to be amortized across tasks.
Second, we present \sys, an architecture that automatically constructs and maintains this memory and uses it to synthesize and execute complete task programs with a single planning call, falling back to reactive reasoning when necessary.
Third, across 655 WebArena tasks, we show that reusing application knowledge improves success rate while substantially reducing latency and cost, and characterize search-dependent interactions as an important boundary where reactive execution remains useful.
Our results show that reactive reasoning need not be the default for every action: a web agent can learn how an application works, reuse that knowledge across tasks, and react only when the environment demands it.

\section{Related Work}
\label{sec:related_work}
\minihead{Reactive Web Agents}  Most existing web agents follow a reactive execution paradigm:
the agent observes the current state, invokes an LLM to select the next action, executes it, and repeats until task completion.
These approaches differ in how they perceive and interact with the application.
Vision-based agents such as ScreenAgent~\cite{niu2024screenagent} and SeeClick~\cite{cheng2024seeclick} reason directly over screenshots, providing a general interface that closely resembles human GUI interaction.
Text-based agents such as WebWISE~\cite{tao2023webwise} and
MindAct~\cite{deng2023mind2web} instead operate over structured webpage representations such as HTML or DOM trees.
Multimodal agents like WebVoyager~\cite{he2024webvoyager} combine visual and textual information to make more robust decisions. 
AgentOccam~\cite{yang2025agentoccam} demonstrates that a simple reactive agent can achieve strong performance without task-specific strategies or additional agent modules by carefully designing its observation and action spaces.

Despite these differences, these agents share the same step-by-step execution loop: each action requires observing the result of the previous action and reasoning about what to do next.
As task trajectories grow longer, this repeated inference increases both latency and cost.
Other work improves step-wise reasoning by incorporating world models to enhance agent foresight.
For instance, WebDreamer~\cite{gu2024webdreamer} simulates the outcomes of potential actions before execution to select the optimal step, while~\cite{gur2023real} explores model-based planning for web navigation. 
Such approaches provide greater foresight when choosing actions, but still interleave LLM reasoning with execution at each step.
\sys instead investigates whether the repeated planning loop can be replaced by one-shot planning followed by programmatic execution.

\minihead{Memory for Web Agents} Several web and GUI agents create and maintain a memory as they execute tasks.
For example, agents like Synapse~\cite{zheng2023synapse} use a simpler form of memory by retrieving flat, past action sequences ("Trajectory-as-Exemplar") as context. 
Other work has explored graph-based memory representations. PageGraph~\cite{chen2025pg} stores page transition relationships in an offline graph structure.
AutoDroid~\cite{wen2024autodroid} represents mobile apps as GUI transition graphs and uses LLM-generated functional summaries to annotate GUI elements. AutoDroid focuses on optimizing accuracy through richer GUI annotations (e.g., injecting task descriptions into HTML 'onclick' properties).
These approaches differ from \sys along two dimensions.
First, memory is used primarily as context for reactive action selection and the agent still reasons about and executes one step at a time.
\sys instead uses memory to synthesize a complete executable program before execution.
Second, existing graph memories primarily capture GUI navigation and page transitions.
The SMG represents reusable application operations, including both operations that transition between states and operations that access data without changing the application state.
This richer operational representation allows the Planner to reason about both control flow and intermediate data when composing a program.



\section{State-Machine Memory}
\label{sec:crawling_agent}

To plan a complete task before execution, the agent needs knowledge beyond its current observation: what application states exist, what operations are available in each state, what information those operations return, and where they lead. 
\sys represents this knowledge as a state-machine graph (SMG).
This representation separates the application's finite set of interface templates from its potentially unbounded dynamic content, keeping the memory compact for planning.

\subsection{Representing an Application as a State Machine}
\label{sec:smg}

\sys represents a web application as a directed graph \(M=(S,O,T)\): states \(S\) are stable application views (\eg a forum's \texttt{Home} and \texttt{Forum} pages), operations \(O\) represent reusable  capabilities available in those states (\eg \texttt{GoToKthPost}, \texttt{ReadPosts}), and the transition function \(T\) records how operations change state.

\minihead{States as interface templates}
A state represents a stable GUI view, such as a webpage.
Each state consists of \emph{atoms}, which group related GUI elements that appear together, such as the buttons in a navigation bar or the fields describing a post.
Atoms may be shared across states (\eg navigation bar appears on many pages), but each state is uniquely defined by
its specific combination of atoms.

\begin{figure*}[t]
    \centering
    \includegraphics[width=\textwidth]{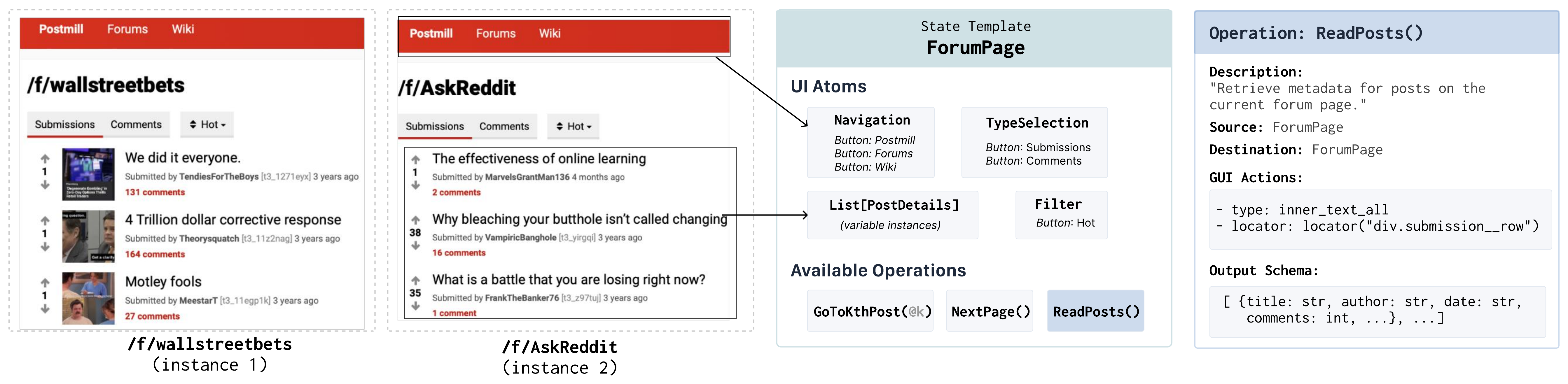} 
\vspace{-1.5em}
\caption{State-machine representation of a web application.
Pages with the same interface structure map to a shared state template, which consists of reusable UI atoms and exposes operations over them.
Operations store how to access live application data rather than the data observed during construction; for example, \texttt{ReadPosts} specifies how to retrieve posts and their output schema.
}
\vspace{-1em}
\label{fig:state-op}
\end{figure*}

Importantly, states capture recurring application structure rather than individual page instances.
As \autoref{fig:state-op} illustrates, \texttt{/f/wallstreetbets} and \texttt{/f/AskReddit} contain different posts but correspond to the same \texttt{ForumPage} state because they share the same interface structure. 
The state is identified by the atom signature
\{\texttt{Navigation}, \texttt{TypeSelection}, \texttt{Filter}, \texttt{List}$[$\texttt{PostDetails}$]$\}.
Here, \texttt{List[PostDetails]} specifies that the page may contain zero or more instances of the \texttt{PostDetails} atom; the number and contents of these instances do not affect the identity of the state.
Consequently, \emph{the graph grows with the number of distinct interface templates rather than the amount of content displayed by the application}.
Appendix~\ref{appendix:smg} describes the atom representation and how \sys resolves state equivalence in more detail.

\minihead{Operations as reusable capabilities}
Interactions with a state are exposed as \emph{operations}. 
An operation is a functional unit composed of one or more primitive GUI \emph{actions}, such as clicking, typing, or reading an element.
Each operation specifies its inputs, source and destination states, the primitive GUI actions that implement it, and output schema if the operation returns data.
For example, \autoref{fig:state-op} shows \texttt{ReadPosts} operation, which reads the current post instances and returns their metadata according to a fixed schema.

Operations form SMG’s edges. Navigation and update operations may transition to another state, while data-collection operations generally produce self-loops because they read information without changing the current view. Data-access operations additionally specify the schema of their outputs, which can be bound to variables and used by subsequent operations during execution.

Crucially, for dynamic content the SMG stores \emph{access paths}, \emph{i.e.,} the mechanisms for retrieving live data, rather than the particular data observed during construction. For example, \texttt{ReadPosts} stores the GUI actions needed to locate and read posts together with its \texttt{List[PostDetails]} output schema; it does not store specific posts observed during crawling. 
Executing this operation later retrieves the current contents of the application. Thus, the same materialized graph can be reused across tasks even as the application's data changes.

\subsection{Automated SMG Construction}
\label{sec:graph_construction}

We assume black-box access to the target application: \sys can interact with the application through its user interface, but does not have access to its source code or an existing model of its structure. 
The Crawling Agent constructs the SMG automatically through exploration of the live application. 
Starting from a seed state, it discovers reachable interface states, identifies the operations available in each state, and records their implementations, inputs, outputs, and transitions. It then executes the discovered operations against the live application to validate them and enrich their descriptions with behavioral details that are not apparent from the interface alone, such as pagination behavior or valid form options. Appendix~\ref{appendix:graph_construction} describes the construction procedure in detail. Although our implementation focuses on web applications, this construction procedure extends naturally to other graphical interfaces (such as mobile and desktop apps).

Importantly, initial SMG construction is \emph{task-independent}: the Crawling Agent observes only the application and its interactions, not the tasks that the Execution Agent will later solve. The resulting graph therefore captures reusable application knowledge rather than solutions or shortcuts for particular tasks. We serialize this knowledge as YAML containing the states, operations, actions, locators, descriptions, inputs, outputs, and transitions used by the Execution Agent.
Figure~\ref{fig:sketch}(b--c) shows an excerpt of this representation, including a UI operation (\texttt{GoToKthPost}) and its referenced UI atom (\texttt{PostDetails}).

Because the Crawling Agent is LLM-driven and black-box exploration may not exercise every application behavior, the initial SMG can be incomplete.
For example, sparse application data or insufficient interaction coverage may prevent the crawler from activating a control that reveals a hidden dialog.
\sys therefore supports subsequent \emph{SMG refinement}: when execution exposes missing or incorrect application knowledge, the \patch can repair the corresponding states or operations and commit the repairs back to the SMG for reuse by subsequent tasks (Section~\ref{sec:patch}).

\section{Online Planning and Execution}
\label{sec:execution_agent}

At runtime, the Execution Agent uses the SMG to turn a user task into a program and execute it with minimal step-by-step reasoning.
As shown in \autoref{fig:architecture}, this process involves three components:
the \emph{Planner} generates an executable program based on the task and the SMG; the \emph{Executor} runs this program deterministically; and the \emph{Patcher} reactively handles failures and patches the SMG.

\subsection{Planner: From Tasks to Executable Plans}
Given a user task and the SMG, the Planner constructs a complete program before execution rather than deciding one action at a time.
The planner separates the what from the how: 
it first uses a single LLM call to generate a high-level \emph{sketch program} over operations in the SMG, then deterministically links these operations through the SMG and compiles them into an executable plan.

\lstdefinestyle{ListingSketchStyle}{
    style=PyStyle,
    keywordstyle=[2]\color{violet}\bfseries,
    keywordstyle=[3]\color{black}\bfseries,
    morekeywords=[2]{OP},
    morekeywords=[3]{execute_task,
                     classify_posts,
                     find_forum_index,
                     LLMCall,
                     ExtractJsonFromLLMResponse},
}

\begin{figure*}[t]
\centering
\begin{minipage}[t]{0.35\textwidth}
\vspace{0pt}
\begin{lstlisting}[style=ListingSketchStyle]
async def execute_task():
    OP: NavigateToForum(
        @name="Showerthoughts")
    OP: SortPostsByNew()
    OP: GoToKthPost(@k=0)
    OP: NavigateToComments()
    vote_counts = OP: ReadVoteCounts()
    count = sum(1 for v in 
      vote_counts if int(v) < 0)
    return count
\end{lstlisting}
\centering
\small (a) Sketch program
\end{minipage}
\hfill
%
\begin{minipage}[t]{0.33\textwidth}
\vspace{0pt}
\begin{lstlisting}[style=YAMLStyle]
name: GoToKthPost
description: 
  "Go to Kth post on the 
  current forum page."
src_state: ForumPage
dst_state: PostDetailPage
actions:
  - input: ['@k']
    action: PostDetails.
            PostLink.click
\end{lstlisting}
\centering
\small (b) SMG Operation 
\end{minipage}
\hfill
%
\begin{minipage}[t]{0.3\textwidth}
\vspace{0pt}
\begin{lstlisting}[style=YAMLStyle]
name: PostDetails
ui_elements:
- name: PostLink
  actions:
  - type: click
    input: '@k'
    locator: locator(
      "nav.submission__nav"
      ).nth(${k})
...
\end{lstlisting}
\centering
\small (c) Referenced UI Atom
\end{minipage}

\caption{
(a) The Planner generates a sketch that expresses application
interactions as abstract \texttt{OP} calls while using ordinary Python
for task logic.
(b) Each \texttt{OP} refers to an operation in the SMG, which specifies
its source and destination states and references actions on UI atoms.
(c) The corresponding atom stores the primitive GUI action and locator
used to execute the operation.}
\vspace{-1em}
\label{fig:sketch}
\end{figure*}

\minihead{Sketch generation} \autoref{fig:sketch} illustrates this process for the task ``Count the comments with more downvotes than upvotes made by the user who posted most recently in the Showerthoughts forum.''
The sketch is ordinary Python, except that interactions with the application are expressed as abstract \texttt{OP} calls.
For example, the Planner generates \texttt{GoToKthPost(@k=0)} rather than the low-level GUI actions needed to open the first post.
The sketch can freely combine these operations with ordinary Python control and data flow: in this example, the vote counts returned by \texttt{ReadVoteCounts} are processed locally to compute the final answer.

Operation arguments are represented as \emph{symbolic inputs}, denoted by an \texttt{@} prefix.
Their values may be specified directly in the sketch or bound at runtime to values produced earlier in the program.
For example, a sketch may extract a username from the current page and later invoke \texttt{GoToUser(@name=username)}, where \texttt{username} is resolved from the execution context.
For tasks requiring semantic reasoning over runtime values, the sketch may additionally invoke an LLM as an explicit helper function.

To generate the sketch, the Planner calls a coding model with the user task, the relevant SMG operations (including their names, descriptions, inputs, and output schema), and a system prompt containing code generation rules and guidance, and explanations related to the structure of the SMG.
The model is also provided an LLM-call interface that the generated
program can invoke for execution-time semantic reasoning.
Importantly, the Planner sees the interface exposed by SMG operations, but not their primitive GUI implementations.
The initial planning call therefore determines the program logic and structure even when parts of its computation depend on runtime values.

\minihead{Planning optimizations: State Filtering and Prefix Caching}
For large applications, the SMG may contain thousands of operations, while an individual task may require only a dozen. 
Exposing the full set of operations to the Planner is therefore both costly and unnecessary. W
e observe that the number of states is much smaller, so \sys uses states as a coarse-grained index into the SMG: a lightweight LLM call first selects a small set of task-relevant states, after which the Planner is given the operations reachable from those states. 
This reduces both the filtering space and the resulting planning context, without requiring the filter to reason over individual operations.

State filtering creates a challenge for prefix caching because different tasks retrieve different portions of the SMG, producing different planning-prompt prefixes.
To recover cache reuse, we introduce a \emph{layered prefix-caching} algorithm that groups related states into a small number of clusters.
The planning prompt places globally shared content first, followed by the operations from the best-matching cluster and then any remaining task-specific operations.
Tasks that access similar regions of the SMG can therefore share most of
their prompt prefix even when their selected states differ.
Appendix~\ref{sec:execution-agent-impl} describes both optimizations in
detail.

\minihead{Linking and Compilation} The sketch specifies the operations needed to solve the task, but may omit intermediate operations needed to connect them.
The linker uses the source and destination states of each operation to identify such gaps, and uses breadth-first search (BFS) over the SMG to find and insert connecting paths.
For loops, it additionally connects the end of one iteration to the start of the next.

Once linked, the program contains a valid sequence of SMG operations but still refers to the application through abstract \texttt{OP} calls.
The compiler deterministically replaces each operation with its implementation stored in the SMG.
As shown in \autoref{fig:sketch}(b--c), \texttt{GoToKthPost(@k=0)} resolves to \texttt{PostDetails.PostLink.click}, whose atom definition contains the primitive \texttt{click} action and GUI locator.
Symbolic inputs such as \texttt{@k} are preserved during compilation rather than resolved immediately, since their values may depend on runtime data.

\subsection{Executor: Deterministic Program Execution}
The compiled program is represented as a control-flow graph (CFG), with GUI actions and Python computations isolated into individual executable nodes.
GUI-action nodes are translated into the corresponding Playwright calls, while Python nodes implement the computation and control flow specified by the sketch.

The \exec executes this plan node-by-node without further planning calls.
It follows the plan's control flow in a sandboxed environment using a restricted Python interpreter that exposes only whitelisted built-ins (\eg \texttt{math} and \texttt{json}) and blocks unsafe system calls such as file and network access.
To support data flow across nodes, the \exec maintains an execution context that tracks variable scope and values produced during execution.
When a GUI action contains a symbolic input, the Executor resolves it against this context before issuing the corresponding Playwright call.

Executing at node granularity also provides precise failure localization: a runtime error can be attributed to the individual GUI action or Python computation that produced it.
This allows failures to be handed to the Patcher without discarding successful execution of the rest of the plan.

\subsection{Patcher: Reactive Failure Recovery}
\label{sec:patch}
When execution fails, the \patch reactively diagnoses and repairs the failure rather than regenerating the entire plan by default.
Because failures are localized to individual execution nodes, the Patcher can localize the failure and resume execution from the repaired region.

For code-related failures (\eg \texttt{JSONDecodeError} or incorrect
Python logic), the Patcher uses the stack trace and execution context to edit the faulty code block.
For GUI-action failures (\eg a timeout caused by a stale locator), it can inspect the current application with the GUI automation tool and re-ground the failed action.
If the failure reveals stale or incorrect knowledge in the SMG (\eg wrong locator, poor operator description), the Patcher also commits the repair back to the SMG so that subsequent tasks reuse the corrected operation.
In contrast, task-specific plan repairs remain local to the current execution.
More substantial planning errors can trigger local plan edits or, when necessary, regeneration of a larger portion of the plan.
After repair, execution resumes from the affected region.

\section{Evaluation}
\label{sec:evaluation}

\subsection{Experimental Setup}

\minihead{Benchmark and methods}
We evaluate \sys on four domains from WebArena~\cite{zhou2023webarena}:
Postmill, Shopping, Shopping Admin, and GitLab.
retrieval and aggregation, form-based workflows, and repository management.
We compare against two strong reactive agents:
\emph{AgentOccam (AOccam)}~\cite{yang2025agentoccam}, a published web agent that uses vision-based observations, and \emph{Claude Code with Playwright MCP}, which interacts with websites through programmatic browser commands and direct DOM access.
Both agents plan and act step-by-step without a persistent model of the environment.
All methods use Claude Opus 4.6 as the backbone model.

For \sys, we first construct an initial SMG using the offline crawler and then refine it during a separate warm-up phase.
During warm-up, we generate and execute one task for each WebArena task template and use the \patch to update the SMG when execution exposes missing or incorrect application knowledge.
Warm-up uses the task templates but not the evaluation task instances or their instantiated parameters.
The \patch is disabled during evaluation, so each reported task reflects a single execution attempt \emph{without online repair}.
Unless otherwise noted, \sys uses the refined SMG and our layered prefix caching optimization.
We analyze both in Section~\ref{sec:ablations}.


\minihead{Evaluation protocol}
We report task success rate using WebArena's evaluator, end-to-end latency, and LLM API cost computed from input and output token usage.
WebArena's reference-based evaluator can reject correct answers because of formatting differences or multiple valid answers, a known limitation~\cite{thakkarwebarena,lu2025agentrewardbench,zhu2025establishing}.
We therefore manually review failed tasks for all methods.
We count objectively correct answers that differ only in formatting or from the reference answer as successful; for subjective tasks, we retain the original evaluator judgment.
We apply the same correction criteria to all methods.

\begin{table*}[t]
\centering
\small
\caption{Performance on WebArena by domain.
All systems use Claude Opus 4.6.
Overall, \sys improves success rate by 8.5 percentage points over the strongest reactive baseline while reducing average latency by $3.2\times$.  }
\label{tab:main_results}
\setlength{\tabcolsep}{3pt}
\begin{tabular}{@{}ll  ccc ccc@{}}
\toprule
& & \multicolumn{3}{c}{\textbf{Success Rate (\%) $\uparrow$}}
& \multicolumn{3}{c}{\textbf{Avg. Latency (s) $\downarrow$}} \\
\cmidrule(lr){3-5} \cmidrule(lr){6-8}

\textbf{Domain} & \textbf{\# Tasks} & \textbf{AOccam} & \textbf{Claude} & \textbf{\sys}
& \textbf{AOccam} & \textbf{Claude}  & \textbf{\sys} \\
\midrule

Reddit  & 106         & 76.2 & 92.4 & \textbf{95.0}
                     & 536  & 59  & \textbf{18} \\
Shopping & 187    & 42.7 & 79.7  & \textbf{87.0}
                     & 363  & 97  & \textbf{28} \\
Shop. Admin & 182 & 58.4 & 74.2  & \textbf{95.3}
                     & 1665 & 101 & \textbf{31} \\
GitLab & 180   & 58.3 & 89.0  & \textbf{89.0}
                     & 499  & 79   & \textbf{28} \\
\midrule
Overall & 655   & 56.8 & 82.7  & \textbf{91.2}
                     & 787  & 87   & \textbf{27} \\
\bottomrule
\end{tabular}
\end{table*}

\subsection{End-to-End Performance}
\label{sec:main-results}

\begin{figure*}[t]
\centering
\includegraphics[width=\linewidth]{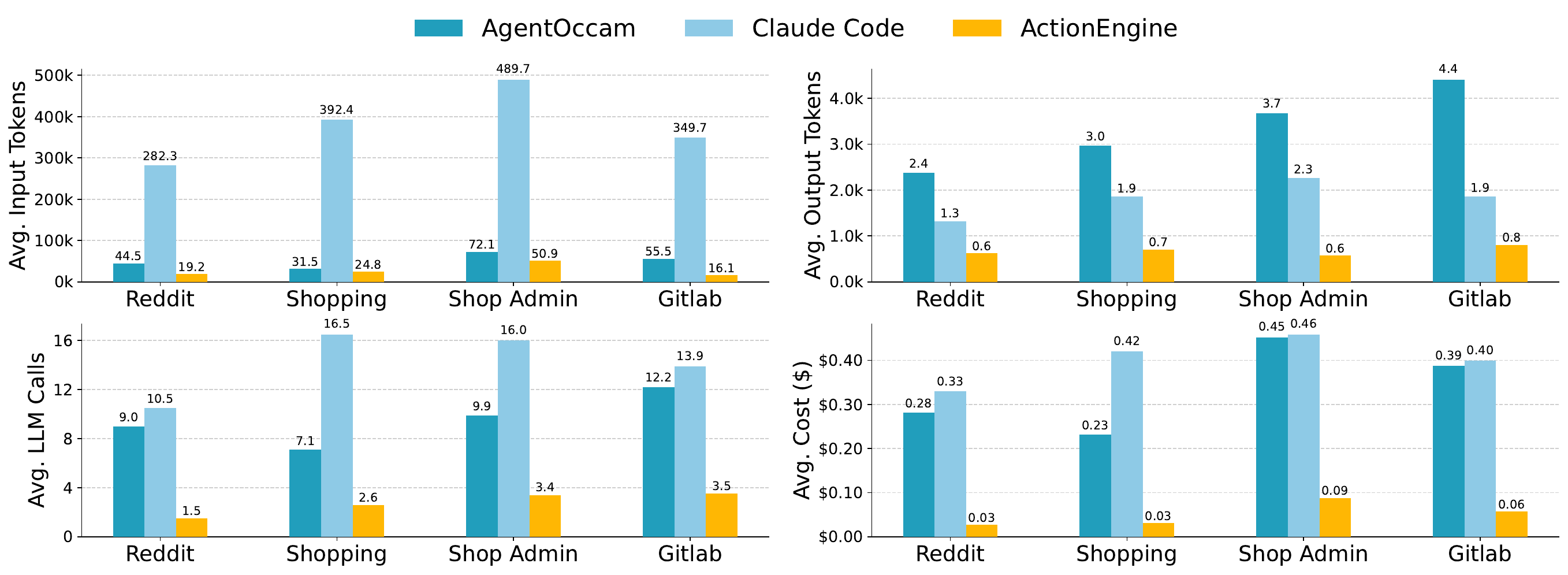}
\vspace{-1.5em}
\caption{Average input tokens, output tokens, LLM calls, and estimated API cost per task.
\sys reduces input tokens by $14$--$22\times$ and LLM calls by $4$--$6\times$ compared to Claude Code. }
\vspace{-1.5em}
\label{fig:efficiency}
\end{figure*}

\autoref{tab:main_results} summarizes success rate and latency across all four domains; 
\autoref{fig:efficiency} breaks down average token usage and estimated cost per task.

\minihead{Success rate} \sys achieves 91.2\% overall success, outperforming Claude Code (82.7\%) by 8.5 percentage points and AgentOccam (56.8\%) by 34.4 points using the same backbone model.
The gain over Claude Code is modest on Reddit (95.0\% vs.\ 92.4\%), but grows to 7.3 points on Shopping and 21.1 points on Shopping Admin, which contain more complex multi-step form workflows.
On GitLab, \sys matches Claude Code at 89.0\%.


\minihead{Latency} \sys reduces average task latency to 27~seconds, a $29\times$ speedup over AgentOccam (787s) and $3.2\times$ over Claude Code (87s).
This improvement follows from replacing iterative reactive execution, where each step requires a separate LLM call and browser round-trip, with one-shot plan generation followed by deterministic execution.
The reduction in the number of LLM calls also reflects this: from 7--16 calls per task for reactive baselines to 1.5--3.5 for \sys (\autoref{fig:efficiency}).
Calls beyond the initial planning call arise from two sources:
a generated program may explicitly invoke an LLM helper for semantic reasoning that depends on values available only at runtime, and large SMGs may require an additional lightweight state-filtering call before planning (Appendix~\ref{sec:sketch-phase}).
%
The speedup is most dramatic on Shopping Admin, where \sys completes tasks in 31s compared to 1{,}665s for AgentOccam---a $54\times$ reduction---as this domain's deep multi-page workflows incur the highest per-step overhead under reactive execution.

\minihead{Cost} Token costs are estimated using Anthropic API pricing (\$5/M input tokens, \$25/M output  tokens) and token usage metrics from the Claude Agent SDK~\cite{agentcost}.
AgentOccam incurs an average cost of \$0.34 per task. Although Claude Code consumes  significantly more input tokens (282k--490k per task), its effective cost (\$0.40) is  not proportional to raw token counts because it leverages prompt caching, where repeated context prefixes are served at 10\% of the standard rate.

\sys reduces average input tokens by $14$--$22\times$ compared to Claude Code and $2\times$ compared to AgentOccam, and output tokens by $2$--$4\times$ and $4$--$6\times$, respectively.
Together with its prefix-caching optimization, this reduces the average API cost to \$0.05 per task---$8\times$ lower than Claude Code and $6.8\times$ lower than AgentOccam.


\minihead{Qualitative takeaways}
The SMG is particularly useful for multi-step tasks that require planning across pages or structured extraction from complex interfaces:
the SMG exposes available operations and their outputs upfront, allowing the Planner to compose them without repeatedly rediscovering the application.
After online refinement, none of the observed failures are caused by broken selectors or structural deficiencies in the SMG.
Instead, the remaining failures primarily involve ambiguous task intent (\(\sim\)23 cases) or search-dependent tasks (\(\sim\)22 cases).
The latter expose a key limitation of plan-then-execute agents: the Planner must choose search terms before observing the results and therefore cannot adapt when search behavior is brittle, results are noisy, or the site's terminology differs from the task.
Detailed success and failure cases are provided in Appendix~\ref{sec:detailed-performance}.

\subsection{Detailed Analysis and Ablation Studies}
\label{sec:ablations}

\minihead{SMG construction and refinement}
\autoref{tab:ablation}(a) reports both the one-time cost of constructing an SMG and the quality of the resulting memory.
Across the four applications, crawling takes 36--260 minutes and costs \$11.6--\$37.2.
Because the resulting SMG is reused across tasks, this cost is amortized over subsequent executions.
Relative to Claude Code, \sys's lower per-task execution cost recovers the one-time crawling cost after 39--101 tasks.


The crawler-produced SMGs already support substantial task success.
Using the crawler SMG, \sys achieves 73.1\% overall success rate, on par with Claude Code performance without any online correction except in the Gitlab domain. 
Refinement further improves success across all four domains, with an average improvement of 18.1 percentage points.
The largest improvement appears on Shopping Admin (74.0\% $\to$ 95.3\%) and GitLab (55.0\% $\to$ 89.0\%), both of which have the most complex interaction patterns (multi-level forms, conditional fields, dynamic GUI elements).
Simpler domains like Reddit (89.0\% $\to$ 95.0\%) and Shopping (80.7\% $\to$ 87.0\%) show a smaller gap, as the crawler's initial graph is already close to complete.
Both variants use the same execution pipeline, with the \patch disabled during evaluation, so these differences isolate the quality of the SMG.
The \patch is restricted to repairing general application knowledge and cannot introduce task-specific shortcuts or composite operations, so refinement improves reusable application knowledge rather than memorizing solutions to evaluation tasks.

We also note that these gains require little change to the structure of the SMG.
Refinement does not add any states, and changes the number of operations only modestly: Reddit and Shopping each gain two operations, Shopping Admin loses four through merging redundant operations that are confusing to the planner, and GitLab gains 66.
Most improvements instead come from repairing or enriching existing
operations.
Using Shopping Admin as a representative example, these repairs fall into three categories:
\emph{(1)~Selector and locator repairs}: the crawler sometimes produces CSS selectors that do not trigger the intended JavaScript event handlers (e.g., flyout menu expansion); the \patch replaces these with Playwright role-based locators that behave correctly.
\emph{(2)~Timing fixes}: actions that trigger asynchronous GUI updates (e.g., AJAX grid refreshes after applying a filter, or modal dialog animations) require explicit waits before the next operation; the \patch inserts these based on observed failures.
\emph{(3)~Description enrichment}: the \patch augments operation descriptions with application behavioral details discovered through execution---for example, noting that a save action stays on the edit page rather than returning to the grid, or documenting how a site's search engine handles fuzzy vs.\ exact matching.




\begin{table*}[t]
\centering
\caption{
(a) Cost and quality of the SMG produced by offline crawling.
Parenthesized operation counts report operations gained or lost after refinement; refinement does not change the number of states.
Break-even is the number of tasks required to amortize the one-time crawling cost relative to Claude Code.
(b) Effect of prefix caching on average API cost per task.}
\label{tab:ablation}

\begin{subtable}[t]{0.68\textwidth}
\centering
\caption{SMG construction and refinement}
\small
\setlength{\tabcolsep}{2pt}

\begin{tabular}{lccc|ccc}
\toprule
\textbf{Domain}
& \textbf{Cost}
& \textbf{Time}
& \textbf{Break-even}
& \textbf{States}
& \textbf{Ops}
& \textbf{Succ.} \\
&
\textbf{(\$)}
& \textbf{(min)}
& \textbf{(\# Tasks)}
&
&
& \textbf{(\%)} \\
\midrule

Reddit
& 11.6
& 36
& 39
& 57
& 130 (+2)
& 89.0 \\

Shopping
& 16.9
& 132
& 44
& 59
& 105 (+2)
& 80.7 \\

Shop Admin
& 37.2
& 260
& 101
& 184
& 1,182 (-4)
& 74.0 \\

GitLab
& 29.5
& 230
& 88
& 219
& 1,148 (+66)
& 55.0 \\

\bottomrule
\end{tabular}
\end{subtable}%
\hspace{0.02\textwidth}%
\begin{subtable}[t]{0.29\textwidth}
\centering
\caption{Prefix caching}
\small
\setlength{\tabcolsep}{3pt}

\begin{tabular}{lrr}
\toprule
\textbf{Domain}
& \textbf{w/o Cache}
& \textbf{Cache} \\
\midrule
Reddit
& \$0.11 & \$0.03 \\

Shopping
& \$0.14 & \$0.03 \\

Shop Admin
& \$0.26 & \$0.09 \\

GitLab
& \$0.10 & \$0.06 \\
\midrule
Overall
& \$0.16 & \$0.05 \\
\bottomrule
\end{tabular}

\end{subtable}

\end{table*}



\minihead{Prefix caching}
\autoref{tab:ablation}(b) isolates the effect of our layered prefix-caching optimization.
Without caching, \sys costs \$0.16 per task on average; caching reduces this to \$0.05, a $3.2\times$ reduction.
The savings are largest on Reddit and Shopping, where the SMG is small enough to expose all operations directly to the Planner and tasks therefore share most of the planning prompt.
Caching remains effective on the larger Shopping Admin and GitLab domains, reducing cost from \$0.26 to \$0.09 and from \$0.10 to \$0.06, respectively.
In these domains, state filtering makes the SMG context task-dependent, reducing the prefix naturally shared across tasks; layered prefix caching recovers reuse by organizing related states into shared prompt prefixes.



\section{Conclusion}
In this work, we introduced \sys, an architecture that transitions Web agents from reactive, step-by-step interaction to programmatic execution.
By capturing reusable application knowledge in an updatable state-machine memory, \sys synthesizes complete executable programs in a single planning step and executes them largely without further reasoning.
On WebArena, this approach improves both task success and efficiency, achieving 91.2\% success while reducing latency by $3.2\times$ and cost by $8\times$ relative to Claude Code.

\bibliographystyle{ACM-Reference-Format}
\bibliography{main} 

\appendix

\section{Offline Memory Construction}
\label{appendix:crawling_agent}

The Crawling Agent simulates a user trying to ``learn'' the GUI application -- \emph{What is this application useful for and how do I use it?}
Rather than saving and trying to reuse task-specific solutions~\cite{zheng2023synapse, zhang2025symbiotic}, the Crawling Agent tries to learn the entire topology of the GUI application (the distinct pages/windows and their constituent GUI elements) and the meaning of every GUI element (e.g., clicking on this next button will navigate to the next page of posts under this forum). 
It then organizes this knowledge as a state machine graph (SMG) memory, which is an intuitive representation that developers already use to design GUIs.

The Crawling Agent constructs the SMG offline through systematic exploration, capturing the intrinsic structure of the application independent of any downstream task.
The design explicitly decouples the static topology (bounded by a finite set of templates) from dynamic data content (potentially unbounded) to keep the state space compact and reusable.
The resulting compact, symbolic representation serves as the foundation for the Execution Agent's one-shot planning capabilities at runtime.

\subsection{State Machine Graph}
\label{appendix:smg}


The state machine graph represents a GUI application as a directed graph whose nodes correspond to symbolic states and whose edges correspond to executable operations. Formally, we define the graph as a state machine $M = (S, O, T)$, where $S$ is a set of discrete states, $O$ is a set of executable operations, and $T : S \times O \rightarrow S$ is the transition function. 

\begin{figure}[!t]
    \centering
    \begin{subfigure}[c]{0.48\columnwidth}
        \centering
        \includegraphics[width=\linewidth]{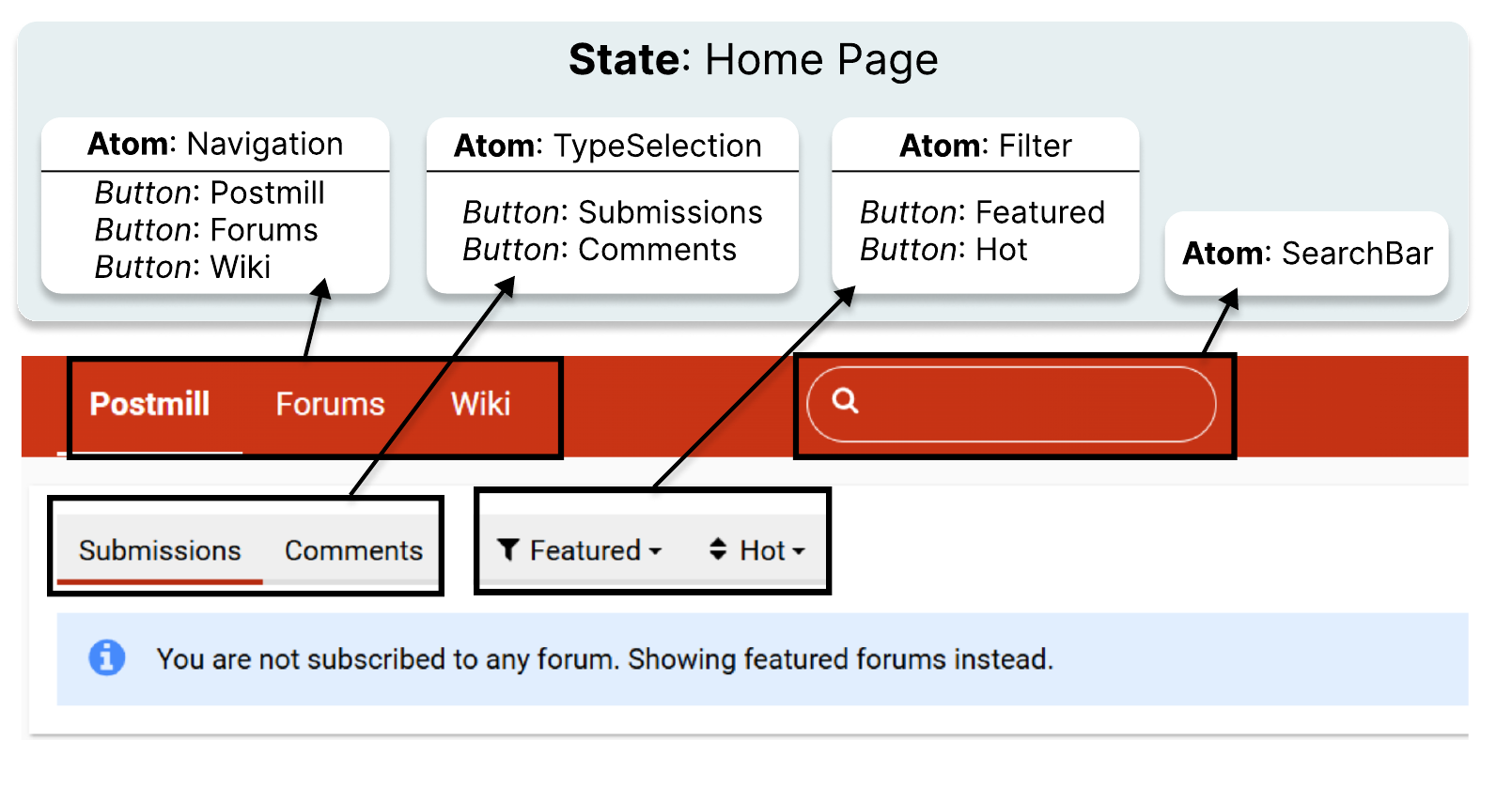}
        \caption{Atoms in the \texttt{Home Page} state.}
        \label{fig:atoms-actions}
    \end{subfigure}
    \hfill
    \begin{subfigure}[c]{0.48\columnwidth}
        \centering
        \includegraphics[width=\linewidth]{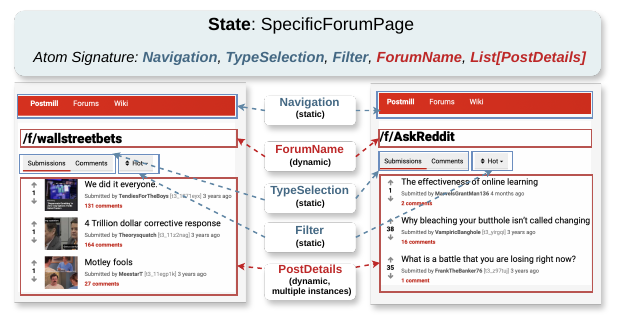}
        \caption{State equivalence across dynamic page instances.}
        \label{fig:static-dynamic-atoms}
    \end{subfigure}
    \vspace{-0.5em}
    \caption{Application states and their Atoms.}
    \label{fig:atoms-dynamic}
    \vspace{-1em}
\end{figure}

A \textbf{State} $s \in S$ represents a unique and stable \emph{view} of a GUI container (e.g., a webpage, a window) for GUI elements (e.g., buttons, menus, tabs, text fields). It is formally defined as a collection of atoms, where an \textbf{Atom} represents a set of related GUI elements that appear atomically (i.e., either all of the GUI elements in an atom are present in the view or none of them). Although atoms may be shared between states, a state is \emph{uniquely} identified by the \emph{combination} of its atoms. 
%
For example, the \texttt{HomePage} state in \autoref{fig:atoms-actions}, is characterized by four atoms: the \texttt{Navigation} atom containing three buttons (Postmill, Forums, Wiki), the \texttt{SearchBar} atom containing a text input and a button, the \texttt{PostTypeSelection} atom containing two buttons (Submissions and Comments) and the \texttt{Filter} atom containing two selection lists. The \texttt{Navigation} and \texttt{SearchBar} atoms are shared between all the states of the Postmill/Reddit application.

Interactions with the GUI application are modeled similarly using a two-level hierarchy of Actions and Operations. An \textbf{Action} captures a \emph{primitive} interaction with a \emph{single} GUI element in an atom (e.g., click on a button, type in a text field, or move the scroll bar). An \textbf{Operation} $op \in O$ is a functional, goal-oriented unit composed of one or more low-level actions. For example, the operation \texttt{Sort Posts By New} includes two actions: \texttt{click(`Sort Dropdown')} followed by \texttt{click(`New Option')}.

\begin{sloppypar}
An operation is represented as an edge in the SMG. 
It operates on atoms pertaining to a \emph{single} state and may result in a \emph{single} state transition. 
It is possible to define \emph{compounded operations} with multiple state transitions (spanning multiple edges in the SMG), but in the current implementation we only experimented with single state operations.
The functionality of these operations vary from simple navigation (e.g., \texttt{NextForumPage}, \texttt{GoToForumList}, etc.), to 
reading or updating content (\texttt{ReadAllPosts}, \texttt{ReadKthPost(k)}, \texttt{UpdateUserBio(new\_bio)}).
GUI manipulation operations typically induce a transition to a new state (e.g., \texttt{GoTo ForumList} from \texttt{HomePage} leads to the \texttt{ForumListPage} state), whereas Data-collection operations introduce self-loops because they do not change the agent's state in the graph (e.g., \texttt{ReadAllVoteCounts} on the \texttt{PostDetails} page). Instead of triggering a state transition, these operations extract information from the GUI and store it in symbolic variables that can be referenced in the online execution stage.
\end{sloppypar}


\minihead{Preventing State Explosion} A central design challenge in the construction of the SMG is preventing state explosion.
For example, a large application such as Amazon contains billions of distinct items customers can inspect, yet modeling each item-specific page as a separate state would make the state graph intractable.
To remain usable for planning, the graph size should scale with the number of distinct \emph{state templates}, not with the number of \emph{data instances} rendered within those templates. 

\autoref{fig:static-dynamic-atoms} illustrates this key property of our state abstraction: a single state template may have an unbounded number of instantiations, each populated with different data. 
In this example, the two forum pages for ``wallstreetbets'' and ``AskReddit'' contain different posts, but they correspond to the same \texttt{ForumPage} state because they share an identical page structure.
This state equivalence is achieved by distinguishing \emph{static} and \emph{dynamic} atoms. 
Static atoms encode invariant interface elements (\eg \texttt{Navigation}),
while dynamic atoms represent elements with data-dependent content (\eg \texttt{ForumName} and \texttt{PostDetails}).
Each dynamic atom essentially defines a ``type'' (analogous to a class definition in object-oriented programming) that can be instantiated zero or more times with different values.

\begin{sloppypar}                
Under this formulation, the \texttt{ForumPage} template is defined by its atom signature: \{\texttt{Navigation}, \texttt{TypeSelection}, \texttt{Filter}, \texttt{ForumName}, \texttt{List}$[$\texttt{PostDetails}$]$\}. The typed collection \texttt{List}$[$\texttt{PostDetails}$]$ specifies that the page expects zero or more instances of the $\texttt{PostDetails}$ atom. 
Although the number of and the content of these instances differ across pages, the state identity is determined by this fixed structural signature, not by the specific data rendered within it.
\end{sloppypar}

\subsection{SMG Construction}
\label{appendix:graph_construction}
The SMG is constructed offline through a fully automated crawl-and-validate pipeline. Starting from a seed state (e.g., the home page), 
the Crawling Agent incrementally builds a complete and consistent representation 
of the application's interaction structure through four phases:

\emph{(1)~State discovery.}
The agent systematically explores the application with the help of a GUI automation tool to enumerate distinct states and transitions. 
It performs a depth-first traversal of the interface, identifying and classifying outgoing links on each view into: (i)~navigation links, explored exhaustively; 
(ii)~data-item links, where representative instances are sampled; and 
(iii)~action-triggering controls (e.g., buttons that reveal hidden forms or wizards).
This phase yields an initial set of states and transitions capturing the application’s reachable topology.



\emph{(2)~Operation definition.}
For each discovered state, the agent defines GUI operations (actions that manipulate the GUI) and data operations (read-only extractions).
Each operation is associated with a concrete grounding (e.g., selectors or locators).
Data operations additionally record an expected output schema (e.g., a list of strings or a flat array) so the Execution Agent knows the format of returned data and can correctly incorporate it into downstream logic.
%


\emph{(3)~Description enrichment.}
The agent probes the live site and enriches each operation's description with behavioral details that cannot 
be inferred from the GUI structure alone.
For example, a search operation's description may specify whether matching is exact, fuzzy, or prefix-based; a pagination operation 
may record the default sort order and items per page; a form operation may list valid dropdown options and index conventions.
These details are critical because the Planner relies solely on operation descriptions to understand what each operation does and how to invoke it correctly.

\emph{(4)~Validation.} Each operation is tested against the live site: GUI operations verify that selectors 
resolve to existing elements; data operations execute the read and validate both the selector and output format. Failures trigger a fix-and-retest loop until all operations pass.

The current design treats the application as a black box, relying solely on live UI 
interaction to discover functionality; inspecting application source code, when 
available, could supplement crawling to uncover hard-to-reach or conditionally 
rendered states.

\minihead{GUI-driven design} Our design philosophy is to ensure that the memory graph captures the intrinsic structure of the GUI application, rather than overfitting to the specific tasks.
Operations are therefore defined purely based on observable GUI elements and their available interactions, independent of any task semantics.
We explicitly avoid introducing ``virtual'' operations that encode composite or task-specific intent.
For example, if a post has three clickable entry points (Title, Thumbnail, ``Read More''), each is recorded as a distinct operation inducing a parallel transition, even if they lead to the same destination state.
The Execution Agent is responsible for reasoning over these alternatives during planning. 

While the initial SMG enforces a strict one-to-one mapping between operations and GUI elements, it can be extended over time based on usage.
In particular, frequently repeated sequences of operations (e.g., \texttt{Login} $\rightarrow$ \texttt{Open Settings}) can be identified and cached as reusable \emph{compounded operations}, similar to materialized views.
This reduces the planning overhead by allowing the agent to reuse common interaction patterns instead of rediscovering them each time.

\minihead{Serialization Format} The finalized graph is serialized into a YAML format optimized for the Execution Agent. \autoref{lst:reddit-state-graph} shows an excerpt from the Reddit SMG, illustrating two representative operations: \texttt{GoToKthPost}, which transitions from \texttt{ForumPage} to the \texttt{PostDetails} state, and \texttt{ReadPosts}, which returns a list of strings with information about each post displayed on the current forum page (e.g., self loop on the \texttt{ForumPage} state). The locators and action types for every Atom's GUI element, which uniquely identify the GUI element and the possible interactions, are specific to the GUI automation tool used, in our case Playwright. While operations' inputs and outputs are not strictly typed, the Planner reliably infers how to parse and use them from the natural-language descriptions and example output formats provided in the SMG. 

\begin{lstlisting}[float, caption={Excerpt from the YAML encoding of the Reddit state-machine graph, 
illustrating a UI operation (\texttt{GoToKthPost}) and a data operation (\texttt{ReadPosts}).},label={lst:reddit-state-graph},style=YAMLStyle]
# ===== STATES =====
- !State
 name: ForumPage
 atoms:
   - name: Navigation
   - name: TypeSelection
   - name: ForumName
   - name: PostDetails
     type: List
# ===== ATOMS =====
- !Atom
  name: PostDetails
  type: dynamic
  ui_elements:
    - name: "PostLink"
      actions:
        - type: click
          input: '@k'
          locator:
            locator("nav.submission__nav").nth(${k})
    - name: "PostInfo"
      actions:
        - type: inner_text
          input: '@k'
          locator: locator("div.submission__row").nth(${k})
        - type: inner_text_all
          locator: locator("div.submission__row")
# ===== OPERATIONS =====
- !Operation
  name: GoToKthPost
  description: "Navigate to the Kth post on the current forum page. Index starts from 0."
  src_state: &ForumPage
  dst_state: &PostDetailPage
  actions:
    - input: ['@k']
      action: PostDetails.PostLink.click
- !Operation
  name: ReadPosts
  description: "Retrieve titles and metadata for posts on the current forum page."
  src_state: &ForumPage
  dst_state: &ForumPage # self-loop
  actions:
    - action: PostDetails.PostInfo.inner_text_all
  output_format: "['Post book depression\n\nSubmitted by bertiewoooster t3_124f4e2 3 years ago\n\n185 comments\n965', ...] - ['Title\nPost Author\nDate\nNumber comments\nNumber Votes', ...]"

\end{lstlisting}

\section{Online Planning and Execution}
\label{appendix:execution_agent}

With the SMG in place, providing a global understanding of the GUI app, the Execution Agent's \emph{\plan} is now able to
create an \emph{executable plan} for the task with \emph{a single reasoning step} (one call to the model). The executable plan is essentially the control flow graph (CFG) of a (Python) program, except that basic blocks may be further split such that each action (GUI interaction) is isolated in its own basic block. The Execution Agent's \emph{\exec} then executes this plan one basic block at a time, for better error isolation and efficient repair. Any error encountered while executing the plan or failure to produce an executable plan (e.g., due to missing operations in the SMG, or poor descriptions) causes the Execution Agent to fallback on the reactive approach: the error/failure is presented to the Execution Agent's \emph{\patch}, which operates in an observe--reason--act loop until the issue is resolved (by either simply editing the code in the faulty basic block, or changing the plan, or updating the SMG).

\subsection{Generating a Sketch Program}
\label{sec:sketch-phase}

The first step towards an executable plan is to generate a \emph{sketch program}: a (Python) script that specifies the task’s logical structure (e.g., control and data flow) while abstracting concrete GUI interactions (actions) into operation placeholders. The Planner achieves this with a call to a coding model, which receives as context: a) the (relevant) operations from the SMG (their names, unique ids, descriptions, any inputs, and any outputs), b) a carefully engineered system prompt, which contains code generation rules and guidance, and explanations related to the structure of the SMG, c) \emph{LLM call as a tool}, for tasks that require reasoning at execution time (e.g., determining if a post recommends a single book), and d) the user task.

The sketch program serves as a \emph{logically complete} blueprint of the task. Providing operations as context (omitting the actions), allows the coding model to focus on the higher-level solution, not on a full implementation (which will be realized through the next deterministic stages). The operation placeholders use the format
\texttt{OP: [OpID]: OpName}. 
The \texttt{OpID} uniquely identifies an operation in the SMG, avoiding ambiguity when multiple operations share similar names. Data-collection operations are tagged with metadata describing their output format, which enables the coding model to incorporate retrieved data into program logic.

The sketch program supports \emph{symbolic variables}, denoted by an \texttt{@} prefix (e.g., \texttt{@username}), 
to represent values that are only available at runtime.
For example, a data-collection operation may first extract raw text into a local variable (e.g., \texttt{comments}); a Python Node—a locally executed code block (e.g., for-loops, parsing, filtering, aggregation, and other data-processing logic)—then processes \texttt{comments} to derive a target value (e.g., the manager's name). 
The Planner specifies how this derived value should bind to symbolic arguments \texttt{@username} in downstream GUI operations. 

Finally, the sketch program also supports \emph{helper functions}, specialized (Python) functions that encapsulate complex logic (e.g., LLM-based inference, multi-step parsing) separate from the main GUI-driven control flow, improving modularity when tasks require sophisticated data processing.


\lstdefinestyle{ListingSketchStyle}{
    style=PyStyle,
    keywordstyle=[2]\color{violet}\bfseries,
    keywordstyle=[3]\color{black}\bfseries,
    morekeywords=[2]{OP, @k},
    morekeywords=[3]{execute_task,
                     classify_posts,
                     find_forum_index,
                     LLMCall,
                     ExtractJsonFromLLMResponse},
}

\begin{lstlisting}[caption={Example sketch program with OP placeholders, symbolic variables, and helper functions.}, 
    style=ListingSketchStyle, label={lst:sketch}]
def find_forum_index(forums_data, name):
    # returns index of forum matching name, or None
    ...

async def classify_posts(post_details):
    prompt = f"""For each post, determine if it recommends exactly one specific book.
    Posts: {post_details}
    Return JSON: {{"urls": ["url1", "url2"]}}"""
    parsed_json = await LLMCall(prompt)
    return parsed_json["urls"]

async def execute_task():
    OP: [1]:GoToForums()
    forums_data = OP: [15]:ReadForums()
    idx = find_forum_index(forums_data, "books")
    while idx is None:
        has_next = OP: [16]:GoToNextPageOfForums()
        if has_next == "0":
            break
        forums_data = OP: [15]:ReadForums()
        idx = find_forum_index(forums_data, "books")
    OP: [6]:GoToKthForum(@k=@idx)
    OP: [12]:SortPostsByTop()
    post_details = []
    while len(post_details) < 10:
        page_posts = OP: [19]:ReadPosts()
        for i in range(len(page_posts)):
            OP: [29]:GoToKthPost(@k=@i)
            title = OP: [21]:ReadPostTitle()
            body = OP: [22]:ReadPostBody()
            url = OP: [0]:ReadURL()
            post_details.append({"title": title, "body": 
                body, "url": url})
            OP: [4]:GoBackToPreviousPage()
            if len(post_details) >= 10:
                break
        has_more = OP: [17]:GoToNextPageOfPosts()
        if has_more == "0":
            break
    result = await classify_posts(post_details)
\end{lstlisting}

\minihead{Optimization: State Filtering} For small applications, all operations are included directly in the planning prompt.
Larger applications, however, can expose hundreds of states and thousands of operations---far more than a single task requires. 
Including the full set wastes 
tokens and introduces irrelevant context that can degrade plan quality.
In such cases, we invoke a \emph{state filter}: a lightweight LLM call that takes the full list of state names and descriptions along with the user task, and returns a small subset of relevant states (typically 3--6). 
Only the operations associated with these states are then passed to the Planner. 
Filtering on states rather than operations keeps the filter prompt compact, as the number of states is often at least an order of magnitude smaller than the number of operations.
State filtering is triggered when the total operation count exceeds 300; below this threshold the overhead is unnecessary. 
Since state filtering produces variable prompt 
subsets across tasks, we design a layered prefix caching scheme to enable efficient cache reuse under this variable-prefix setting (\S~\ref{sec:execution-agent-impl}).

\minihead{Example (Sketch Program)} Consider the example task: \textit{Among the top 10 posts in "books" forum, show me the post URLs that recommend a single book.} 
The \plan generates the sketch program shown in \autoref{lst:sketch} (simplified for clarity).
The sketch demonstrates several key features: (1) \texttt{OP} placeholders with operation IDs (e.g., \texttt{[16]}, \texttt{[19]}) that will be expanded to concrete actions after compilation; (2) the helper function \texttt{classify\_posts} encapsulates LLM-based reasoning as a specialized Python function, keeping complex inference logic separate from the main GUI-driven control flow; (3) symbolic variables like \texttt{@idx} whose values are computed at runtime and bound dynamically; (4) control flow (\texttt{if-else}) is fixed at plan time based on the task structure, while the actual branch taken depends on runtime data (e.g., collecting enough posts).



\subsection{Compiling into Executable Plans}
\label{sec:connection-phase}

The \plan transforms the abstract sketch into a fully executable plan in two steps: \emph{linking}, which resolves each \texttt{OP} placeholder into concrete execution paths in the SMG, and \emph{compilation}, which assembles the result into an execution-ready plan with runtime symbolic binding.
Both steps are performed programmatically and deterministically via a custom compiler without any model calls.

First, the \plan expands each \texttt{OP} placeholder into a valid sequence of graph edges in the SMG while preserving the control-flow semantics established in the IR.
Before expansion, it validates connectivity: for each consecutive pair of operations, a breadth-first search (BFS) checks whether a valid transition exists from the previous operation’s end state to the next operation’s start state.
For operations inside loops, the check additionally verifies that the path returns to the loop’s entry state after each iteration.
If any disconnection is detected, the specific break points are reported to the Planner's model, which applies a targeted local edit to repair the affected portion of the sketch.
In our evaluation with Claude Opus 4.6, generated sketches were always connected and this repair step was never triggered; however, it serves as a safety net for weaker planning models that may produce inconsistent operation sequences.

\begin{sloppypar}
Once all \texttt{OP}s have been expanded, the system compiles the resulting program into an executable plan that is type-safe, execution-ready, and supports runtime symbolic binding. 
The sketch's control-flow graph (CFG) is constructed with each 
action isolated into its own basic block (CFG node).
Each \texttt{OP} is then replaced with its corresponding implementation (e.g., Playwright API call).
For example, $\langle$\texttt{ACTION} type=\texttt{click}, locator=\texttt{locator("nav.submission\_\_nav")\allowbreak.nth(@i)}$\rangle$ will be replaced with the \texttt{page.locator("nav\allowbreak.submission\_\_nav")\allowbreak.nth(i).click()} Playwright call, ready for execution. Each basic block carries its execution context, code logic, and data binding requirements.
\end{sloppypar}

\section{Implementation}
\label{sec:implementation}


\subsection{Crawling Agent}
\label{sec:crawler-impl}

The Crawling Agent is implemented as a set of skills on top of Claude Code with the Playwright MCP server, which provides programmatic browser control via the Model Context Protocol.
The agent reads page structure through Playwright's accessibility snapshots and interacts with elements using standard Playwright selectors and locators.

The crawling pipeline combines LLM-driven exploration with hard-coded validation scripts.
The LLM handles tasks requiring semantic understanding, such as identifying page templates, classifying interactive elements, and composing operation descriptions. 
Deterministic scripts enforce convergence: phases 1 and 2 are each gated by a validation script that must pass before the next phase begins, with failures triggering a fix-and-retest loop.
For example, the phase~2 convergence check performs a BFS from the seed state and enumerates all interactive and data-bearing DOM elements on each reachable page. It flags two types of gaps: elements with no corresponding operation, and elements where an existing operation's selector fails to match. The agent iterates---adding 
missing operations or repairing selectors---until the script reports full coverage.
Phase 4 is itself a validation pass that executes every operation on the live site; failures are fed back to the LLM for repair until all checks pass.

For the most complex application in our evaluation (GitLab, 103 states, 1{,}082 operations), the full four-phase pipeline completes in approximately 1.5 hours, consumes 704k input tokens, and costs approximately \$5---a one-time offline 
investment amortized across all subsequent task executions.

\subsection{Execution Agent}
\label{sec:execution-agent-impl}



\label{sec:patcher-impl}
We implemented the Planner and Executor with 11,017 lines of Python code.
The \plan produces a \texttt{MixedActionPlan}---a tree of typed nodes that the 
\exec traverses at runtime.
Each \texttt{OP} expands into its SMG action sequence; 
Python lines aggregate into Python block nodes; and control-flow statements are 
converted via AST analysis into recursive container nodes preserving the original 
nesting. Helper functions are hoisted to a root-level node so they are defined before 
the main execution flow.

The \patch is implemented as a Claude Agent SDK agent equipped with Playwright MCP as a built-in tool, and read/write access to the SMG.
The runtime invokes the \patch asynchronously when a failure occurs, passing it a detailed execution trace including visited URLs, executed actions, and their success/failure status.
The failure handling logic is described in \S~\ref{sec:execution-phase}; SMG edits made by the \patch are persisted so that subsequent task executions automatically benefit from the fix.

\minihead{Optimization: Layered Prefix Caching} The state filter (§\ref{sec:sketch-phase}) selects task-relevant states, but different tasks select different subsets, yielding distinct prompt prefixes that prevent cache reuse.
We introduce a three-layer prompt structure that restores a shared, cacheable prefix across tasks.

GUI apps expose \emph{global navigation operations} (e.g., \texttt{go\allowbreak\_to\_dashboard}) available from every state.
These create a star topology in the state transition graph that prevents meaningful partitioning.
We first extract global operations into a shared layer, then partition the residual graph into $K$ balanced clusters (Algorithm~\ref{alg:op-clustering}).
Louvain community detection~\cite{blondel2008fast} splits large connected components along natural business-domain boundaries, so that states commonly needed together remain in the same community.
An LPT bin-packing step~\cite{coffman1978application} then assigns communities to $K$ size-balanced buckets.
The procedure runs once at 
application load time in $O(|V|{+}|E|)$; on a 143-state application, wall time 
is under 15\,ms.

\begin{algorithm}[ht]
\caption{Clustered Prefix Caching}
\label{alg:op-clustering}
\footnotesize
\KwIn{State-action graph $\mathcal{G}{=}(\mathcal{S}, \mathcal{O}, \mathcal{E})$, clusters $K$}
\KwOut{Global ops $\mathcal{O}_g$, clusters $\{C_1, \dots, C_K\}$, mapping $\phi{:}\,\mathcal{S} {\to} \{1,\dots,K\}$}
\BlankLine
$\mathcal{O}_g \leftarrow \{o \in \mathcal{O} \mid \texttt{src}(o) {=} \varnothing\}$\;
$\mathcal{O}_l \leftarrow \mathcal{O} \setminus \mathcal{O}_g$\;
$H \leftarrow$ undirected graph on $\{\texttt{src}(o) \mid o {\in} \mathcal{O}_l\}$ with edges from $\mathcal{O}_l$\;
\BlankLine
$\bar{n} \leftarrow |\mathcal{O}_l| / K$\;
$\mathcal{P} \leftarrow \varnothing$\;
\ForEach{\textnormal{connected component} $\mathcal{C}$ \textnormal{in} $H$}{
  \eIf{$|\mathrm{ops}(\mathcal{C})| > 1.5\,\bar{n}$ \textbf{and} $|\mathcal{C}| > 3$}{
    $\mathcal{P} \leftarrow \mathcal{P} \cup \textsc{Louvain}(H[\mathcal{C}],\,\lceil|\mathrm{ops}(\mathcal{C})| / \bar{n}\rceil)$\;
  }{
    $\mathcal{P} \leftarrow \mathcal{P} \cup \{\mathcal{C}\}$\;
  }
}
\BlankLine
Sort $\mathcal{P}$ by $|\mathrm{ops}(\cdot)|$ descending\;
Init buckets $C_1, \dots, C_K \leftarrow \varnothing$\;
\ForEach{$P \in \mathcal{P}$}{
  $k^* \leftarrow \arg\min_{k} |\mathrm{ops}(C_k)|$\;
  $C_{k^*} \leftarrow C_{k^*} \cup P$\;
}
\Return $\mathcal{O}_g,\; \{C_1,\dots,C_K\},\; \phi(s){=}k\;\forall s {\in} C_k$
\end{algorithm}

Given the state filter output $\mathcal{S}_{\mathrm{rel}}$, we select the best-matching cluster $k^*$ (covering the most states in $\mathcal{S}_{\mathrm{rel}}$) and assemble the prompt in three layers:
\textbf{Layer~1} (shared across all tasks) contains global operations and system instructions;
\textbf{Layer~2} (shared within cluster $k^*$) contains all operations in the best-matching cluster;
\textbf{Layer~3} (unique per task) contains operations from spillover states not covered by $k^*$ (typically 0--2) plus the task description.
If every cluster is substantially larger than $\mathcal{S}_{\mathrm{rel}}$, Layer~2 is omitted to avoid adding irrelevant context, falling back to a two-layer scheme.
Layers~1 and~2 are each marked with a cache breakpoint, so tasks sharing the same cluster reuse both layers from cache.
Since Layer~2 has only $K$ variants, the theoretical Layer~2 cache hit rate over $N$ tasks is $(N{-}K)/N$ assuming no cache expiration, as each cluster incurs at most one cold miss.

\section{Detailed Task Performance}
\label{sec:detailed-performance}

\subsection{Representative successes}
The majority of tasks complete with a single LLM call; some additionally require lightweight runtime LLM calls for intermediate reasoning (e.g., determining which commenter is the website manager, or selecting a target product from order items).
We highlight four tasks where \sys succeeds but Claude Code fails, illustrating two 
structural advantages of programmatic planning over reactive execution.

\emph{Global planning from complete application knowledge.} 
Because the \plan sees the full SMG upfront, it can design multi-step strategies that span multiple states without iterative runtime discovery or backtracking.
For example, in Task~107 (\emph{``Present the monthly count of successful orders from May to 
December 2022,''} Shopping Admin), the \plan generates a \texttt{for} loop over 
8 months, where each iteration applies a status filter, a date-range filter, reads the record count from a summary operation, and resets filters for the next iteration.
This strategy requires knowing that filter and reset operations exist, that they compose correctly, and that a summary operation can return the count directly without reading individual rows.
Claude Code, seeing only the current snapshot at each step, does not know these operations exist until it discovers them interactively; it forgets to reset filters between months and misidentifies controls whose accessibility labels are auto-generated IDs rather than semantic names.


Similarly, in Task~464 (\emph{``Update the Antonia Racer Tank description with positive user 
reviews,''} Shopping Admin), the \plan designs a three-state workflow: navigate to the Reviews grid to read and filter positive reviews via a runtime LLM call, then to the Products grid to open the product form, then write the composed description.
Because the \plan sees operations across all states simultaneously, it plans the optimal data-gathering-then-writing workflow upfront.
Claude Code navigates to the product page first, realizes it needs review content, backtracks to the Reviews page, then returns---but the grid retains the previous search filter, preventing it from finding the target product.

\emph{Grounded operations enabling efficient data extraction.}
Because each SMG operation carries validated selectors and known output schemas, 
\sys extracts data cleanly without parsing noisy page snapshots.
In Task~321 (\emph{``How much refund from orders cancelled in 2022, including 
shipping?''} Shopping), the \plan reads each table column (ID, date, status, total) via separate data operations with precise selectors, producing aligned lists that a helper function zips and filters.
Claude Code must extract the same fields from full-page accessibility snapshots at each pagination step, parsing $\sim$2{,}000 lines mixing headers, data rows, and navigation elements, leading to row-column misalignment and an incorrect total.

Task~284 (\emph{``Show the least expensive shoe storage with minimum 12-pair 
capacity,''} Shopping) illustrates how the two advantages work together to generate a more efficient execution strategy. 
Knowing that a single data operation can return all product names and prices at once, the \plan reads the full list in one shot and passes it to a single batch runtime LLM call that evaluates all candidates simultaneously. Claude Code, unaware of this operation, clicks into each product individually (one reasoning step per product), misses candidates, and reports an incorrect result.


\subsection{Failure analysis} 
Notably, after online SMG refinement, zero failures are caused by selector or graph structural issues. The system also correctly identifies 7 tasks that are impossible on the target site (e.g., \emph{``Change the delivery address for my most recent order''}, which Magento does not support), reporting them as infeasible rather than producing incorrect actions. 


\emph{Ambiguous tasks} ($\sim$23 cases): the task intent or judgment criteria are ambiguous, and the system produces a reasonable answer that differs from the benchmark reference.
For example, \emph{``Thumbs down the top 2 posts ever in history''} (Reddit) is interpreted as the top posts site-wide rather than posts in the forum named ``history.''
On Shopping, \emph{``How much did I spend on hair care in January 2023?''} requires classifying which products count as ``hair care''---the system includes items like body butter and headbands, yielding a different total than the benchmark.
On Shopping, \emph{``What are the main criticisms of this product?''} requires selecting which review sentences constitute ``main criticisms''---which review sentences constitute ``main criticisms'' is a matter of interpretation.
\emph{``List reviewers mentioning ear cups being small''} (Shopping) finds 4 out of 5 reviewers but misses one who expresses the same complaint with indirect language rather than the exact phrase.
On Shopping Admin, \emph{``5 blue Cronus yoga pants arrived, update the stock''} is ambiguous about whether to add 5 to current stock or set stock to exactly 5.

\emph{Search-dependent tasks} ($\sim$22 failures): the plan-then-execute model struggles when tasks depend on keyword search results that cannot be anticipated at planning time.
For example, \emph{``List all opened issues about OPT model''} (GitLab) searches ``OPT model,'' which returns zero results because the actual issues use titles like ``OPT-175B inference'' and ``Open Pre-trained Transformer''---the site's keyword search cannot bridge this semantic gap.
\emph{``Show the least expensive switch card holder with minimum 15 card capacity''} (Shopping) searches ``switch card holder,'' but the site's search engine returns irrelevant products (phone wallets) alongside the intended Nintendo Switch card holders, and the Planner cannot filter out noise without observing the actual results.
On Reddit, \emph{``Open the thread of a trending post on the forum `machine learning'\,''} fails because the forum is named ``MachineLearning'' ---the exact-string search does not match.

These failures reflect a core limitation of the programmatic paradigm: the \plan must commit to search keywords at plan time without runtime observation. We intentionally disable the \patch during evaluation to stress-test this limitation; with the \patch enabled, \sys can observe search results and iteratively refine keywords, functioning similarly to a reactive agent for search-dependent tasks.

\end{document}